\documentclass[runningheads]{llncs}
\usepackage{graphicx}
\usepackage{amsmath}
\usepackage{amssymb}
\usepackage{hyperref}
\usepackage{subcaption}
\usepackage[export]{adjustbox}[2011/08/13]
\usepackage{makecell}
\usepackage{xcolor, soul}
\sethlcolor{green}
\usepackage{multirow}
\usepackage[misc]{ifsym}
\begin{document}
\title{Towards a multimodal framework for remote sensing image change retrieval and captioning}
\titlerunning{Remote Sensing Image Change Retrieval and Captioning}

\author{Roger Ferrod\inst{1}(\Letter) \and
Luigi Di Caro\inst{1} \and
Dino Ienco\inst{2,3}}
\authorrunning{R. Ferrod et al.}
\institute{University of Turin, Turin, Italy \\
\email{(roger.ferrod,luigi.dicaro)@unito.it} \and
INRAE, UMR TETIS, Univ. Montpellier, Montpellier, France \and
INRIA, Univ. Montpellier, Montpellier, France \\
\email{dino.ienco@inrae.fr}
}
\maketitle              

\begin{abstract}
Recently, there has been increasing interest in multimodal applications that integrate text with other modalities, such as images, audio and video, to facilitate natural language interactions with multimodal AI systems. 
While applications involving standard modalities have been extensively explored, there is still a lack of investigation into specific data modalities such as remote sensing (RS) data. Despite the numerous potential applications of RS data, including environmental protection, disaster monitoring and land planning, available solutions are predominantly focused on specific tasks like classification, captioning and retrieval. These solutions often overlook the unique characteristics of RS data, such as its capability to systematically provide information on the same geographical areas over time. This ability enables continuous monitoring of changes in the underlying landscape.


To address this gap, we propose a novel foundation model for bi-temporal RS image pairs, in the context of change detection analysis, leveraging Contrastive Learning and the LEVIR-CC dataset for both captioning and text-image retrieval. By jointly training a contrastive encoder and captioning decoder, our model add text-image retrieval capabilities, in the context of bi-temporal change detection, while maintaining captioning performances that are comparable to the state of the art. We release the source code and pretrained weights at: \url{https://github.com/rogerferrod/RSICRC}.




\end{abstract}
\section{Introduction}
Modern Earth observation systems allow acquiring systematic information, under the shape of satellite imagery, to monitor and characterize the evolution of the underlying Earth surface. Among all the applications, the possibility to detect and characterized particular changes in the land surfaces is of paramount importance in a variety of applications such as environmental protection, disaster monitoring and land planning \cite{doi:10.1080/10095020.2022.2128902}.


More precisely, the value of such information comes from the detailed comparison between bi-temporal remote sensing imagery and all the related features. 
The majority of current change detection approaches (e.g., \cite{chg2cap,RSICCformer}) exhibit a limited level of interaction from a user perspective. Given a pair of images, they mainly highlight the spatial areas affected by some change phenomena, without any additional information or any strategies to guide the process via possible user query. To express its full potential and allow an appropriate interaction with users, features related to changed areas must be effectively described and searchable, making it possible to interact with the system in natural language. For these reasons, there is a strong interest in the community to develop models that go beyond simple Change Detection (CD) {strategies and try instead to accurately describe bi-temporal changes occurring or retrieve a pair of images associated to a given textual prompt.

Those aims can be achieved through multimodal foundation models (e.g., \cite{Sun2023RingMoAR,Wang2022AdvancingPV,Zhang2023Text2SegRS,remoteclip}) that, once pretrained on a large-scale dataset, can be used in many down-stream tasks with remarkable performance. Foundation vision-language models, led by CLIP \cite{clip}, have already demonstrated excellent capabilities in remote sensing applications, too (\cite{remoteclip,zhang2024earthgpt,hu2023rsgpt}). However, despite great success in other domains, Vision-Language models for Remote Sensing applications still suffer from the scarcity of large-scale datasets. In fact, while there are plenty of works focused on specific tasks -- such as classification \cite{Adegun2023ReviewOD}, captioning \cite{chg2cap} or retrieval \cite{Zhou2023RemoteSI} -- foundational models are  still at their infancy stage in the RS domain. 

Motivated by the lack of Vision-Language models for Remote Sensing applications that manage bi-temporal change detection information, here, we propose a foundation model specifically designed for pairs of bi-temporal remote sensing images. To the best of our knowledge, it is the first attempt to fill this gap in the literature. Given the shortage of resources focused on changes in RS imagery, we propose to  adopt a remote sensing image change captioning dataset (LEVIR-CC) to assess the potential of our framework considering both bi-temporal captioning and bi-temporal text-image retrieval tasks. By jointly training a contrastive encoder and captioning decoder, we provide a single model that, simultaneously, allows bi-temporal text-image retrieval capabilities, preserving captioning abilities comparable to the state of the art.

\section{Related works}
Our work follows a line of research started recently by \cite{RSICCformer} who introduced the task and the dataset we used. The dataset, described more in details in Section \ref{sec:dataset}, is accompanied by a model named RSICCformer. Another model (Chg2Cap) was then proposed by \cite{chg2cap} the following year, obtaining state-of-the-art results on the same benchmark. Both models are limited to the captioning task. Therefore, their efforts are focused on building a solid encoder attached to a simple decoder and train them with a captioning loss. Similar solutions, not limited to the RS domain, include \cite{Chouaf2021CaptioningCI,Hoxha2022ChangeCA,Guo2022CLIP4IDCCF,Yao2022ImageDC,Tu2023NeighborhoodCT,Tu2023ViewpointAdaptiveRD}.

With foundation models, instead, a single pretrained model can be used for many downstream tasks, with only minor fine-tuning and supervision efforts.
Such techniques are now getting more and more attention also in the (RS) domain, with recent advances such as RemoteCLIP \cite{remoteclip}, EarthGPT \cite{zhang2024earthgpt} and RSGPT \cite{hu2023rsgpt}. However, such initiatives are still focused on the analysis of static information and do not take into account the temporal dimension, i.e., the evolution or changes that might occur between two or more remote sensing image acquisitions.

With the aim to combine two different paradigms (retrieval and captioning) into a single model, though, some architectural modifications are required. For this purposed, we based our framework on CoCa \cite{yu2022coca}. In CoCa the authors propose a strategy to train a multimodal foundation model that can perform various tasks, such as captioning, text-image and image-text retrieval and visual recognition. Such work, however, is limited to static, natural images, and it cannot be easily extended to cope with bi-temporal satelletie imagery. 

With regard to the retrieval task, Vision-Language contrastive models like CLIP \cite{clip} and derivatives, such as ALBEF \cite{albef} and ALIGN \cite{align} can be adopted. These multimodal AI models have already contributed to ameliorate state-of-the-art results in many downstream task. Recently, extensions of these models for the RS domain have been proposed, like \cite{AlRahhal2022MultilanguageTF} and RemoteCLIP \cite{remoteclip}, achieving state-of-the-art performance in RS image retrieval task but, unfortunately, such models are still focused on the analysis of static remote sensing imagery, and they cannot be employed for the analysis of bi-temporal remote sensing imagery for downstream change detection applications.

\section{Method}
Building on top of state-of-the-art approaches, we propose a new model to address the challenges posed by retrieving a pair of images related to temporal changes in remote sensing images via natural language prompts. More precisely, our goal is to provide a model to cope with both captioning and text-image retrieval tasks at once, for the case of remote sensing data. To achieve this objective, we integrate a SOTA image-encoder specifically designed for image change captioning with a new decoder that enables a joint training of the two objectives: an autoregressive captioning loss and a contrastive loss. 

Inspired by CoCa \cite{yu2022coca} we split the decoder in two parts: an unimodal module that encodes the textual input only and a multimodal module with cross attention that combines textual and visual embeddings. In both cases, the decoder prohibits tokens from attending to future tokens in the sequence. More details on the decoder are provided in section \ref{sec:decoder}. 

Differently from CoCa, thought, our model needs to deal with a pair of images, not a single one. Having experimented with RSICCformer \cite{RSICCformer} and Chg2Cap \cite{chg2cap} we have decided to adopt Chg2Cap's encoder, with only minor revisions, given its excellent behavior exhibited on the captioning task. The encoder is responsible for encoding the images individually, through a pretrained model, and then combining the features into a single representation. The details related to the encoder are supplied in section \ref{sec:encoder}. The overall architecture is depicted in Figure \ref{fig:arch}.

To test our model, we relied on the LEVIR-CC dataset introduced in \cite{RSICCformer}, which was the first research work to address the Remote Sensing Image Change Captioning (RSICC) task introducing the RSICCformer architecture. Since no datasets are available, today, for remote sensing image retrieval with pairs of images (before change/after change), we exploited the LEVIR-CC dataset both for captioning and retrieval, although this choice required some special precautions discussed in section \ref{sec:dataset}.

\begin{figure}[t]
    \centering
    \includegraphics[width=1\textwidth]{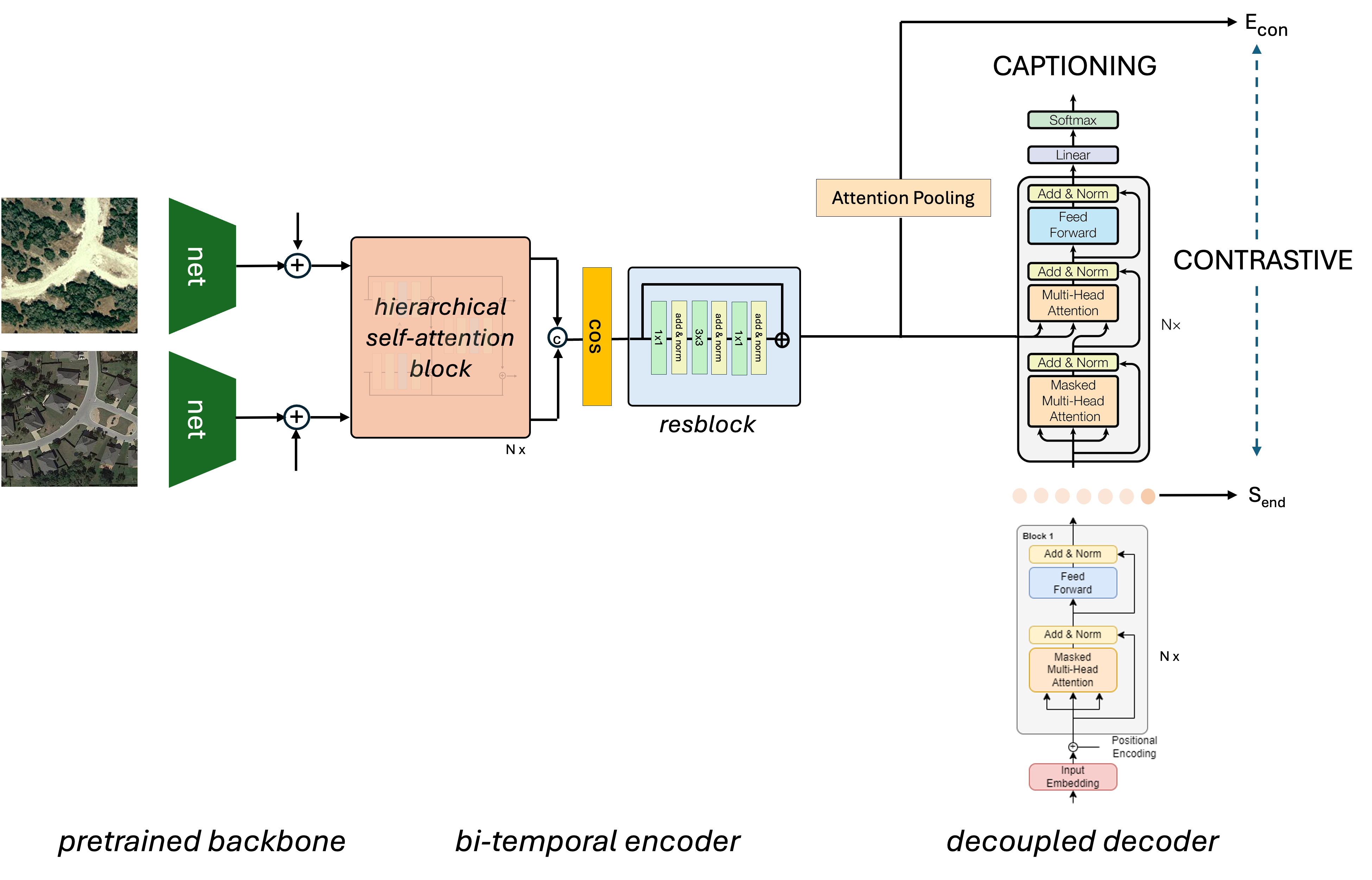}
    \caption{The overall architecture of our model; once the pair of images is encoded through two siamese pretrained models, the information is processed by a bi-temporal encoder that merges the two representations, then a single embedding can be retrieved through attentive pooling and contrastively compared with the corresponding textual embedding or used directly as input for the cross-modal decoder; the decoder is splited in two parts: unimodal layers that only encode the textual representation and multimodal layers that generate the captions.}
    \label{fig:arch}
\end{figure}

\subsection{Dataset}
\label{sec:dataset}
The LEVIR-CC dataset consists of 10,077 image pairs and 5 captions each describing the changes happening between the two images cover the same area  (Figure \ref{fig:dataset}). Half of the pairs do not have any significant change, while the other half portrays changes such as new building or road being built. Often the description includes detailed spatial references (e.g., “on the top-left” or “at the bottom of the scene”) that identify the changes. Although the dataset is vast and has over 50k sentences, the possible changes that could occur are limited by the fact that the acquired images are taken almost in the same geographical zone (Texas, USA), within a similar urban context. Therefore, while it is naturally possible to exploit the data for the captioning task, adapting the dataset for image retrieval is more challenging. Given a textual query, indeed, more images that match the description could be retrieved. Within a contrastive learning framework, those examples are considered false negatives and could make the training stage challenging. 

\begin{figure}[]
    \centering
    \includegraphics[width=.7\textwidth]{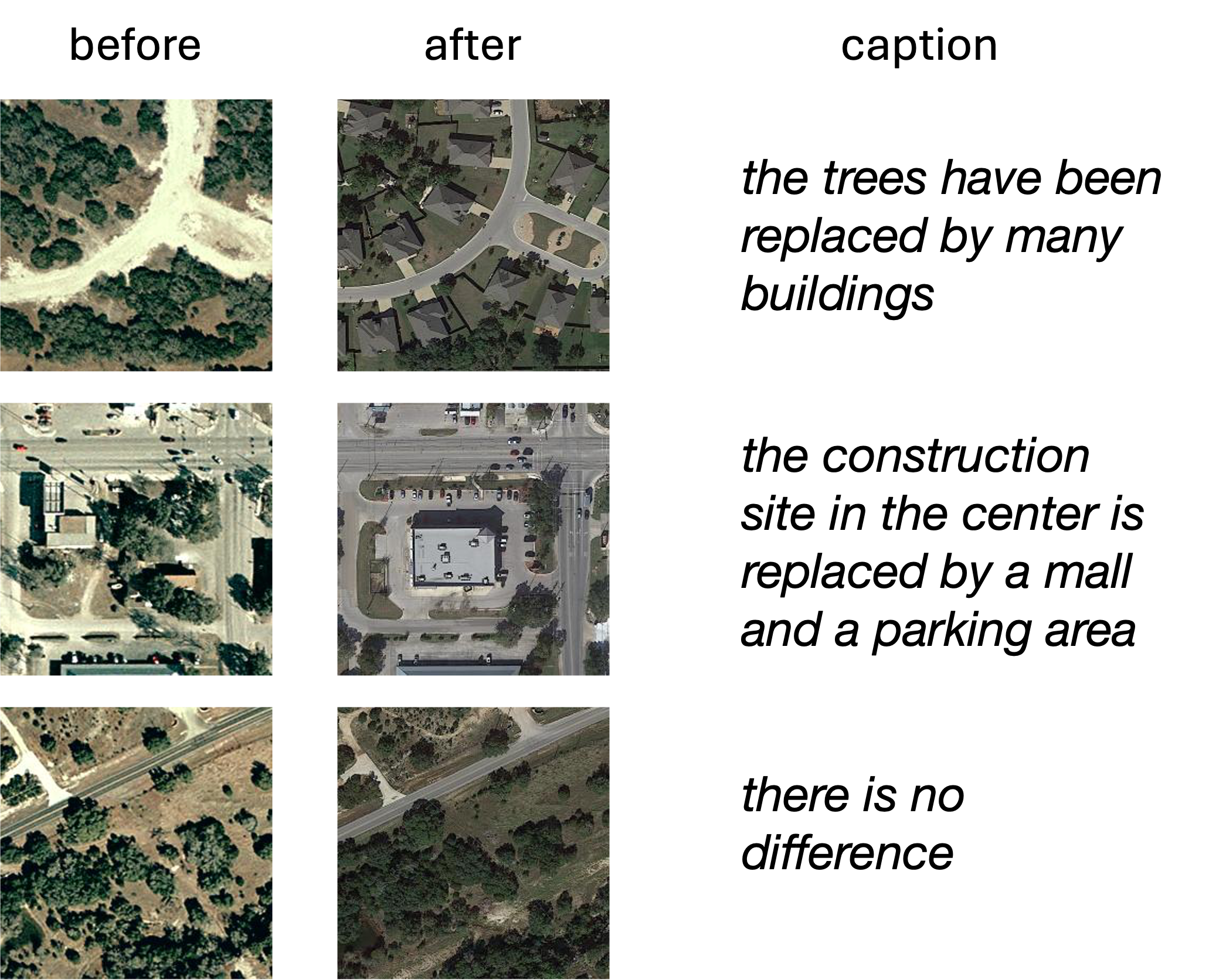}
    \caption{Examples of items taken from the LEVIR-CC dataset, where each image pair (before/after) is accompanied by 5 human annotated captions (only one is shown here).}
    \label{fig:dataset}
\end{figure}

\subsection{Encoder}
\label{sec:encoder}
The encoder follows the same architecture proposed by \cite{chg2cap}, with a pretrained siamese network to encode the image pair, a hierarchical attention mechanism to combine the features and a residual block with cosine mask. Whereas the original Chg2Cap model was proposed with a resnet-101 backbone -- and tested on other models pretrained on ImageNet -- we included in our experiments networks especially pretrained on remote sensing images. If our model is set to fine-tune the encoder's backbone, only the weights from the last two convolutional layers of the ResNet architecture, or the transformer layers of ViT, are updated. 

After a linear projection head used to bring the feature dimension to the desired size, we continue as in the original Chg2Cap model, in particular once we have an embedding representation $F_i$ for each image $X_i$:
\begin{equation}
    \begin{aligned}
        F_1 = net(X_1) \\
        F_2 = net(X_2) 
    \end{aligned}
\end{equation}
with $F_i \in \mathbb{R}^{h \times w \times D}$ and $D$ the feature dimension, we add learnable positional embeddings:
\begin{equation}
    F_i = F_i + F_{pos}
\end{equation}
with $F_{pos} \in \mathbb{R}^{h \times w \times D}$, and pass the extracted features to a hierarchical self-attention block that will apply the attention mechanism across the two images:
\begin{equation}
    I_i, I_2 = HSA(F_1, F_2)
\end{equation}
where $I_i \in \mathbb{R}^{h \times w \times D}$.
When fusing the two representations, through a concatenation over the hidden dimension, a cosine mask is applied as follows:
\begin{equation}
    F_{fus} = [I_1; I_2] + Cos(I_i, I_2)
\end{equation}
where $F_{fus} \in \mathbb{R}^{h \times w \times 2D}$, $[;]$ is the concatenation operation and $Cos(\cdot , \cdot)$ the cosine similarity between the two tensors. 
After applying a 2D convolution layer with 1x1 kernel size and dimensionality reduction, $F_{fus}$ is processed by a residual block (1x1 conv, 3x3 conv, 1x1 conv) obtaining a final encoding ($E_{cap}$) for the image pair:
\begin{equation}
    \begin{aligned}
        C = conv_{1 \times 1}(F_{fus}) \\
        E_{cap} = ReLU(ResBlock(C) + C)
    \end{aligned}
\end{equation}
where $C \in \mathbb{R}^{h \times w \times D}$ and $E_{cap} \in \mathbb{R}^{h \times w \times D}$.

$E_{cap}$ can then be used directly as input for the decoder cross-attention layer to generate a caption. For the contrastive loss used for the text-image retrieval task, instead, a single embedding vector representing the pair of images is derived. To this aim, we added a single query multi-head attention layer as an attention pooling operation:
\begin{equation}
    E_{con} = MHA(E_{cap})
\end{equation}
with $E_{con} \in \mathbb{R}^D$

\subsection{Decoder}
\label{sec:decoder}
The transformer-based decoder requires some modifications to both captioning and contrastive learning. Following CoCa \cite{yu2022coca}, we split the decoder in two parts: the first one will encode the text and can be used directly by the contrastive learning loss, while the second part will combine the textual encoding with visual embeddings and produce the caption. Both apply casually-masked attention to prevent the current token to attend to future tokens, but only the second part will apply cross-attention, combining the textual and visual representations. More formally:
input tokens are mapped into a word embedding using an embedding layer and added to positional embeddings:
\begin{equation}
    E = emb + pos
\end{equation}
where $emb : \mathbb{R}^{n} \rightarrow \mathbb{R}^{n \times D}$, $n$ the number of input tokens, $D$ the embedding dimension and $pos$ the sinusoidal positional encoding. Then the unimodal decoder layers follow, which will produce two results: the self-attention outputs with shape $n \times D$ and a single representation of the sequence taken from the last token:
\begin{equation}
\begin{aligned}
    S_{seq} = MHA(E) \\
    S_{end} = S_{seq}[n]
\end{aligned}
\end{equation}
where $S_{seq} \in \mathbb{R}^{n \times D}$ and $S_{end} \in \mathbb{R}^D$. While $S_{end}$ is used directly by the contrastive loss, $S_{seq}$ will pass through other decoder layers, this time with cross-attention:
\begin{equation}
    E_{cross} = MHA(S_{seq}, E_{cap})
\end{equation}
where $E_{cross} in \mathbb{R}^{n \times D}$, $S_{seq}$ the textual sequence embeddings with shape $n \times D$ and $E_{cap}$ the image embeddings with shape $h \times w \times D$. Finally, a linear layer ($LN$) is used for tokens prediction:
\begin{equation}
    C = LN(E_{cross})
\end{equation}
with $LN \in \mathbb{R}^{n \times |V|}$ where $V$ is the vocabulary. 

Differently from the encoder, which in part exploits pretrained models, the decoder in trained from scratch. The particularity lies in the vocabulary used, that is derived from the dataset itself. Both \cite{RSICCformer} and \cite{chg2cap}, indeed, compose the vocabulary with the words appearing at least 5 times in the dataset, resulting in 463 tokens. Any attempt to go beyond these limits adopting a pretrained decoder with thousands of tokens will drop the performance, making the results not comparable. Therefore, we have decided to follow the same strategy adopted by previous works. We also add the possibility to tie embedding weights with the decoder, resulting in shared parameters. 

\subsection{Training}
As already mentioned, this work aims to combine captioning capabilities with text-image retrieval using only a single model. We achieve this with a modified architecture which allows the two paradigms to coexist. 

\subsubsection{Possible issues related to False Negatives}
However, moving from a single image to a pair of images (before and after a change occurred) is not straightforward and requires a mechanism to tackle false negatives, i.e., examples identified as negative by the contrastive loss, which in fact should be considered as positive. 
This problem arises from using a dataset originally intended for captioning only. The phenomenon is common in many other tasks in which contrastive learning is employed, especially in a self-supervised setting, and solutions exist to cope with this particular issue \cite{9706613,101016,chen2022incremental}. Following the common literature, we implemented False Negative Elimination (FNE) and False Negative Attraction (FNA) strategies. With FNE, the detected false negative is simply removed from the loss computation, while the FNA strategy considers the false negative as a positive example and changes its label accordingly. In our work, the false negative detection phase is done by comparing the captions similarities. If two captions belonging to different image pairs (therefore considered negative examples) have a cosine similarity higher than a given threshold ($\theta$), the FNE or FNA mechanisms are activated. As shown in Figure \ref{fig:fn}, two different captions with similar semantics can then be considered as the same caption (FNA) or excluded from the same batch (FNE). 

\begin{figure}[t]
    \centering
    \includegraphics[width=1\textwidth]{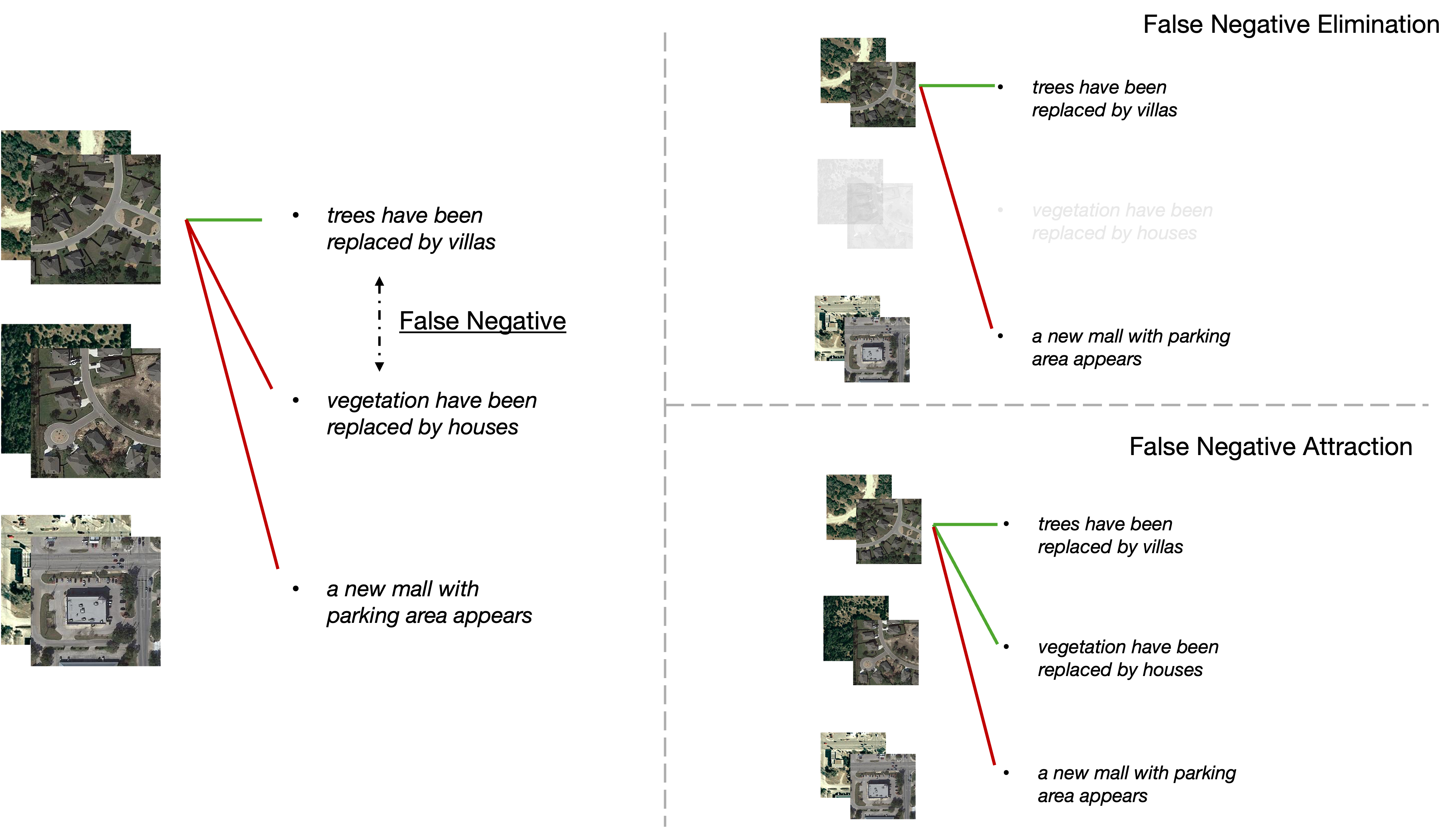}
    \caption{By performing contrastive learning, an anchor (image pair) is compared with the captions inside the batch, the corresponding textual description is considered a positive example, the others negative. If a False Negative is detected (caption similarities higher than $\theta$), one can exclude it from the loss computation (False Negative Elimination) or consider it as positive (False Negative Attraction).}
    \label{fig:fn}
\end{figure}

Since the textual decoder is trained from scratch, for the similarity computation we rely on a third-party model specifically designed for sentence comparison. We pre-compute the embeddings with Sentence-BERT \cite{reimers-2019-sentence-bert} and then the cosine similarity before applying the contrastive loss.

\subsubsection{Contrastive Learning}
To combine the visual representation and textual embeddings, we rely on the InfoNCE loss, a popular contrastive learning loss. Visual and textual encoders are jointly optimized by contrasting pairs of images and captions inside the batch:
\begin{equation}
    \mathcal{L}_{con} = -\frac{1}{N}( \sum_{i}^{N}log\frac{exp(e_i \cdot s_i / \tau)}{\sum_{j}^{N} exp(e_i \cdot s_j / \tau)} + \sum_{i}^{N}log\frac{exp(s_i \cdot e_i / \tau)}{\sum_{j}^{N} exp(s_i \cdot e_j / \tau)} )
\end{equation}
where $N$ is the batch size, $\tau$ the temperature and $e_i$ and $s_j$ the (normalized) visual and textual embeddings, respectively. 

More in detail, the batch is composed by $N$ triples $<e_i, s_i, l_i>$, where $l_i$ is the label of the $(e, s)_i$ pair. We then compute all the possible combinations of contrastive pairs and assign a positive or negative label accordingly to $l_i$, i.e., when $s_i$ is the corresponding caption for the $e_i$ image pair. At this point, we have to correct the labels by detecting and removing false negatives. We take the pre-computed normalized Sentence-BERT embeddings ($t_i$) and compute a $N \times N$ similarity matrix in which we identify the combination of captions -- within the batch -- that have a relative similarity higher than a given threshold $\theta$. Formally, positive and negative examples are defined as: 
\begin{equation}
    (e_i, s_j) = 
    \begin{cases}
    (e_i, s_j)_+ & \textit{if } l_i = l_j \\
    (e_i, s_j)_- & \textit{if } l_i \neq  l_j \land t_i \cdot t_j < \theta
    \end{cases}
\end{equation}
when using False Negative Elimination (FNE), or:
\begin{equation}
    (e_i, s_j) = 
    \begin{cases}
    (e_i, s_j)_+ & \textit{if } l_i = l_j \lor t_i \cdot t_j \geq \theta \\
    (e_i, s_j)_- & \textit{if } l_i \neq  l_j \land t_i \cdot t_j < \theta
    \end{cases}
\end{equation}
when using False Negative Attraction (FNA).

\subsubsection{Captioning}
While the contrastive loss compares the whole representation of the image pair with an analogous representation of the caption, the captioning task  acts at a more fine-grained scale. Indeed, the cross-attention of the decoder considers the entire sequence of tokens and the spatial features from the images. Given this information, the decoder will predict the sequence of tokens autoregressively:
\begin{equation}
    \mathcal{L}_{cap} = -\sum_{i=1}^{n}log P(y_i | y_{<i}, e)
\end{equation}
where $n$ in the sequence length and $P$ the probability of generating the $y_i$ token given the previous words and the image encoding ($e$). 

\subsubsection{Training Objective}
Finally, the model is jointly trained by combining both $\mathcal{L}_{con}$ and  $\mathcal{L}_{cap}$ and weighting the relative contribution in the main loss:
\begin{equation}
    \mathcal{L} = \mathcal{L}_{cap} + \lambda \cdot \mathcal{L}_{con}
\end{equation}

\section{Experiments and Results}
We trained our model on the LEVIR-CC dataset and compared its results with other remote sensing models. Of those, only \cite{RSICCformer} and \cite{chg2cap} deal with a pair of images as input, yet they are focus only on the captioning task. Other remote sensing models, on the other hand, are trained for text-image retrieval and/or captioning, but only on single images. 

\subsection{Setup}
Training and evaluation are performed on a single NVIDIA A40 GPU with 48 GB of memory. We evaluate the model after each epoch, and we select the model that minimizes the loss. The maximum number of epochs is set to 50 and the batch size to 32. The AdamW optimizer is used, with $\epsilon = 10^{-8} $ and linear warm up reaching a target learning rate of $10^{-4}$. The InfoNCE temperature $\tau$ is set to 0.01.

\subsection{Experiments}
We compared our results with the state of the art and other pretrained remote sensing models. 
For those models that were only trained for single-image ranking (\cite{align,remoteclip,albef}) we encoded the image pair as the difference between RGB values, specifically: $E = | X_{after} - X_{before}|$.

We also add the possibility of including hard negatives examples in the batches. A hard negative pair is an example in which the image and textual encodings have a high similarity despite being negative examples, i.e., the text is not the intended caption for that specific image. Therefore, at each epoch we compute and update an index, sampling the negatives examples globally (i.e., across the entire dataset). For this purpose, we rely on the FAISS library~\cite{faiss} and cosine similarity. 

We evaluated the captions with standard text quality metrics like BLEU score \cite{Papineni02bleu} and ROUGE-L \cite{lin-2004-rouge} as well as CIDEr \cite{7299087} for image description evaluation. The same metrics were used in SOTA models like \cite{chg2cap} and \cite{RSICCformer}. For the text-image retrieval task, instead, we relied on Precision (P), Recall (R) and Mean Reciprocal Rank (MRR) over \textit{top-k} results. Specifically, the metrics are defined as follows:
\begin{equation}
    P@k = \frac{\text{\textit{relevant items in top-k results}}}{k}
\end{equation}

\begin{equation}
    R@k = \frac{\text{\textit{relevant items in top-k results}}}{\text{\textit{relevant items}}}
\end{equation}

\begin{equation}
    MRR@k = \frac{1}{Q}\sum_{i=1}^{Q}\frac{1}{rank_i}
\end{equation}
where $Q$ is the number of queries, $rank_i$ the position of the first relevant item in the \textit{top-k} results for the \textit{i-th} query. For each metric, the relevant items are computed by taking into account the $\theta$ threshold used for the false negative detection phase during the training. Therefore, if two examples are merged during the training because their similarity is higher than $\theta$, the same will happen during the evaluation phase. 

\subsection{Results}
When evaluating the results, it is important to consider the uneven distribution of the dataset. Since half of the dataset consists of examples without changes (approximately 5,000 items), a \textit{top-5} search of an unchanged scene would yield a low Recall score (~0.1\%), even if all five matches are relevant.

At the same time, since for each caption we can have only one corresponding image in the original dataset -- or slightly more if we consider the aggregation results induced by the use of the $\theta$ threshold -- for most of the queries the Precision will be near $1/k$, or $20\%$ if $k=5$. 

In Table \ref{tab:results} we compare the baselines with our framework, without contrastive learning, with False Negative Elimination and with False Negative Attraction. 
As highlighted by the obtained results, the contrastive learning strategy allows our model to outperform the baseline approaches. At the same time, we maintained comparable performances on the captioning task, despite having a smaller backbone than the one originally used in Chg2Cap (ResNet-50 instead of ResNet-101). It is also interesting to note that the contrastive learning loss had a positive influence also on the captioning task, with small advantages compared to the baseline without contrastive learning. Among the two strategies, False Negative Attraction came out as the best and confirms the finding already reported in the literature \cite{9706613}. 


\begin{table}[!h]
\centering
\begin{tabular}{ccc|cccc|l}
P@5   & R@5   & MRR@5 & BLEU-1 & BLEU-4 & ROUGE-L & CIDEr & model   \\ \hline 
-  & -  & - & \textbf{82.84}  & \textbf{60.92} & \textbf{72.72} & \textbf{130.97}  & Chg2Cap \\
21.33 & 0.61 & 27.59 & -  & -    & - & -  & ALIGN \\
\underline{41.95} & 0.71 & \underline{52.32} & -  & -    & - & -  & Remote-CLIP RN50 \\
40.74 & 0.28 & 51.85 & -  & -    & - & -  & ALBEF \\\hline
30.1 & 0.52 & 16.87 & 79.55  & 55.76 & 62.15 & 117.14  & w/o contrastive learning \\
41.6 & \underline{1.2} & 52.15 & 69.38  & 37.92 & 55.21 & 106.61  & FNE \\
\textbf{52.32} & \textbf{2.85} & \textbf{53.51} & \underline{82.34}  & \underline{59.04} & \underline{62.99} & \underline{120.14}  & FNA \\
\end{tabular}
\caption{Text-image retrieval and captioning results of SOTA captioning system (Chg2Cap), Retrieval baselines (ALIGN, Remote-CLIP and ALBERF) and our models (without contrastive learning, with False Negative Elimination or False Negative Attraction); the best results are highlighted in bold, the second best are underlined.}
\label{tab:results}
\end{table}

Before choosing the proper threshold value for the false negative detection task, we manually annotated 150 captions randomly sampled from the test set, by merging textual description with the same meaning. Based on these annotations, we can then evaluate the quality of different threshold values, as reported in Table \ref{tab:sim} (a). In particular, with $\theta$ above $0.96$ (in terms of cosine similarity) the false negative detection is perfectly aligned with human judgment, before dropping sharply with lower similarity value. From the Table \ref{tab:sim} (b), instead, we can appreciate the difference between False Negative Elimination and False Negative Attraction compared with a simpler training method that does not apply a false negative detection strategy. We confirm that the False Negative Attraction strategy exhibits the best behavior, when the similarity threshold is set to $1.0$. With this setting, we can achieve better ranking scores and comparable captioning performances compared to the baseline approaches. 

\begin{table}[t]
\centering
\subfloat[]{
\begin{tabular}{cccc|c}
\multicolumn{4}{c|}{}  \\
ACC  & F1   & P & R & $\theta$   \\ \hline 
99.90 & 54.55 & 37.50 & 100 & 0.92 \\
99.92 & 60.00 & 42.86 & 100 & 0.94 \\
99.98 & 85.71 & 75.00 & 100 & 0.95 \\
100 & 100 & 100 & 100 & 0.96 \\
100 & 100 & 100 & 100 & 1.0 
\end{tabular}
}
\quad
\subfloat[]{
\begin{tabular}{c|cc|l}
R@5  & BLEU-1 & BLEU-4 & model   \\ \hline 
2.23 & \textbf{82.64} & \textbf{59.76} & w/o FN detection \\
1.93 & 77.96 & 54.27  & FNE $\theta=0.96$ \\
2.58 & 80.95 & 57.73 & FNA $\theta=0.96$ \\
1.2  & 69.38  & 37.92 & FNE $\theta=1.0$\\
\textbf{2.85} & 82.34  & 59.04 & FNA $\theta=1.0$\\
\end{tabular}
}
\caption{a) evaluation of the threshold over 150 manually annotated samples and b) impact on the model.}
\label{tab:sim}
\end{table}

Although the main difference of our framework compared to previous approaches is the decoder component, we also propose to replace the original backbone used in Chg2Cap (a ResNet-101 pretrained on ImageNet) with a Remote Sensing pretrained model. Given the good results obtained by Remote-CLIP (Table \ref{tab:results}), we have chosen to adopt its encoder as backbone for our framework. Despite the fact that it has fewer parameters than ResNet-101, the pretraining phase on remote sensing images proved to be beneficial, and the model was able to outperform the original encoder pre-trained on ImageNet. In Table \ref{tab:finetune} we compare the fine-tuned ResNet-50 backbone we adopted, with a non fine-tuned version (i.e., with frozen weights) and the same architecture but trained on ImageNet.
Albeit a frozen version of the Remote-CLIP backbone already permits to achieve more than reasonable results on the ranking task, we have observed that fine-tuning even ameliorate the results, as briefly mentioned in Section \ref{sec:encoder}. To this end, we only fine-tune the last two layers.


\begin{table}[]
\centering
\begin{tabular}{c|cc|l}
R@5   & BLEU-1 & BLEU-4 & model   \\ \hline 
1.83 & 73.95 & 45.38 & ImageNet RN50 finetune\\
\textbf{2.9} & 78.47 & 52.47 & RemoteCLIP RN50 no finetune \\
2.85 & \textbf{82.34} & \textbf{59.04} & RemoteCLIP RN50 finetune\\
\end{tabular}
\caption{Comparison between different backbones: RN50 from RemoteCLIP finetuned/non finetuned and the same architecture pretrained on ImageNet.}
\label{tab:finetune}
\end{table}

Concerning the importance of the (global) hard negatives sampling mechanism, no clear evidence is available. Although this solution represents the best practice in Contrastive Learning \cite{Wang2020UnderstandingTB}, it proved to be counterproductive in our case, as shown in Table \ref{tab:hd}. 
Since we rely on self-supervised labels for the ranking task rather than high-quality labeled data, the hard negative mechanism might nullify the effect of false negative detection, thereby hindering the final results.


\begin{table}[]
\centering
\begin{tabular}{c|cc|l}
R@5   & BLEU-1 & BLEU-4 & model   \\ \hline 
\textbf{2.85} & \textbf{82.34}  & \textbf{59.04} & w/o hard negatives \\
0.98 & 72.69 & 43.57 & w/ hard negatives \\
\end{tabular}
\caption{Impact of the hard negative mechanism on performance.}
\label{tab:hd}
\end{table}

\section{Conclusion and Future works}
In this work, we present a new multimodal foundation model designed for bi-temporal remote sensing images, with captioning and text-image retrieval capabilities. Building on top of state-of-the-art solutions, we adapt pretrained visual encoders and a novel multitask decoder with the aim to combining two paradigms into a single model. Given the absence of benchmarks for RS retrieval of bi-temporal images, we propose to exploit a change detection captioning dataset (LEVIR-CC) for both tasks, with some precautions that aim to mitigate the problem of false negatives under contrastive learning training. Although there is still room for improvement in the absolute metric results for the retrieval task, our model maintains captioning performance comparable to the state of the art, with the added benefit of also providing a solution for the text-image retrieval task.

We believe that natural language prompting is crucial to facilitate the exploration of Remote Sensing image archives by non-expert end-users. Our work is a step towards this direction, but there is still work to be done in order to achieve fully reliable solutions. For instance, efforts must be done to create and curate a benchmark for bi-temporal image retrieval, which can be shared with the community to stimulate research activities in this important area. Another possible future research can be devoted to the design and development of generic foundational visual language remote sensing models capable of handling multiple tasks simultaneously, rather than relying on single-task models that require systematic adaptation for each downstream task.

\bibliographystyle{splncs04}
\bibliography{biblio}

\end{document}